\documentclass[a4paper, 10pt, conference, table]{ieeeconf}      

\IEEEoverridecommandlockouts                              

\makeatletter
\let\NAT@parse\undefined
\makeatother

\usepackage{graphicx} 
\usepackage{amsmath}
\usepackage{amssymb}

\usepackage{bm}
\usepackage{bbm}
\usepackage{dsfont}
\usepackage{multirow}
\usepackage{booktabs}
\usepackage{nicefrac}
\usepackage{cite} 	
\usepackage{pdfpages}
\usepackage{filecontents}
\usepackage[utf8]{inputenc}

\usepackage{graphicx}

\usepackage{caption}
\usepackage{subcaption}
\usepackage{refcount}
\usepackage{afterpage}
\newcommand{\setfootnotemark}{\refstepcounter{footnote}\footnotemark[\value{footnote}]}

\usepackage{siunitx}
\usepackage{xcolor}
\usepackage{collcell}
\usepackage{hhline}
\def\colorModel{hsb} 
\newcommand\ColCell[1]{
	\pgfmathparse{#1<50?1:0}  
	\ifnum\pgfmathresult=0\relax\color{white}\fi
	\pgfmathsetmacro\compA{0}      
	\pgfmathsetmacro\compB{#1/100} 
	\pgfmathsetmacro\compC{1}      
	\edef\x{\noexpand\centering\noexpand\cellcolor[\colorModel]{\compA,\compB,\compC}}\x #1
}
\newcolumntype{E}{>{\collectcell\ColCell}m{0.8cm}<{\endcollectcell}}  

\usepackage{tikz}
\usepackage{pgfplots}					
\usepackage{pgfplotstable}
\usetikzlibrary{decorations}
\pgfplotsset{width=7cm,compat=newest, every tick label/.append style={font=\tiny}}	
\usepackage{pgf}

\usepackage[
	colorlinks=true,          
	linkcolor=red!85!blue,    
	citecolor=black!30!green, 
	urlcolor=red!25!blue,     
	breaklinks=true,
	pagebackref=true,         
]{hyperref}
\usepackage[capitalise]{cleveref}

\crefname{table}{Tab.}{Tabs.}
\crefname{section}{Sec.}{Secs.}
\crefname{subsection}{Sec.}{Secs.}

\definecolor{Gray}{rgb}{0.5,0.5,0.5}
\definecolor{darkblue}{rgb}{0,0,0.7}
\definecolor{orange}{rgb}{1,.5,0} 
\definecolor{red}{rgb}{1,0,0} 

\newcommand{\yaw}{\phi}

\newcommand{\radvel}{v_r}

\newcommand{\x}{\Vect{x}}               

\newcommand{\Vect}[1]   {{\ensuremath{\mathbf{\lowercase{#1}}}}} 

\DeclareMathOperator*{\cross-entropy}{cross-entropy}

\title{\LARGE \bf
Using Machine Learning to Detect Ghost Images in Automotive Radar
}

\author{Florian Kraus$^{1}$, Nicolas Scheiner$^{1}$, Werner Ritter$^{1}$, Klaus Dietmayer$^{2}$
\thanks{$^{1}$Mercedes-Benz AG, 70565 Stuttgart, Germany\newline{\tt\small florian.kraus@daimler.com}}
\thanks{$^{2}$Institute of Measurement, Control and Microtechnology, Ulm University, 89081 Ulm, Germany}%
}

\begin{document}

\maketitle
\thispagestyle{plain}
\pagestyle{plain}

\begin{abstract}
Radar sensors are an important part of driver assistance systems and intelligent vehicles due to their robustness against all kinds of adverse conditions, e.g., fog, snow, rain, or even direct sunlight.
This robustness is achieved by a substantially larger wavelength compared to light-based sensors such as cameras or lidars.
As a side effect, many surfaces act like mirrors at this wavelength, resulting in unwanted ghost detections.
In this article, we present a novel approach to detect these ghost objects by applying data-driven machine learning algorithms.
For this purpose, we use a large-scale automotive data set with annotated ghost objects.
We show that we can use a state-of-the-art automotive radar classifier in order to detect ghost objects alongside real objects.
Furthermore, we are able to reduce the amount of false positive detections caused by ghost images in some settings.

\end{abstract}

\section{Introduction}
Advanced driver assistance systems and automated vehicles are major trends in the current automotive industry.
For the environmental perception, a wide sensor suite is used to provide a high robustness and fulfill safety requirements.
The most popular three sensors in this regard are camera, lidar, and radar sensors.
Each sensor has its own advantage, e.g., camera has high angular resolution, lidar has the most dense 3D perception, and radar provides single shot velocity estimation via the Doppler effect.
Another major difference for radar sensors is there substantially larger wavelength around \SI{4}{\milli\meter} for a \SI{77}{\giga\hertz} radar.
This long wavelength allows the radar signal to pass through many objects and conversely makes it much more robust to adverse weather conditions such as snow, rain, and fog.
Nevertheless, a disadvantage for radar waves is that many objects in real world scenarios act like mirror surfaces due to their highly specular reflection properties.
Recently, we could show that mirrored non-line-of-sight detections have the potential to be utilized as an early warning system for collision prevention \cite{Scheiner2020CVPR}.
While applications like this are certainly beneficial towards autonomous driving, the task to detect the presence of a multi-path detection still relies on additional sensors.
In this article, we use a large data set in order to investigate this challenge in real world scenarios.
We utilize machine learning techniques to discriminate mirror objects from real objects and background detections by using only radar data without any additional information.
To this end, we apply PointNet++ \cite{Qi2017}, i.e., an end-to-end neural network architecture which directly processes radar point clouds.
We compare several interesting settings and report quantitative and qualitative classification results.

\begin{figure}
	\centering
	\includegraphics[width=0.99\linewidth]{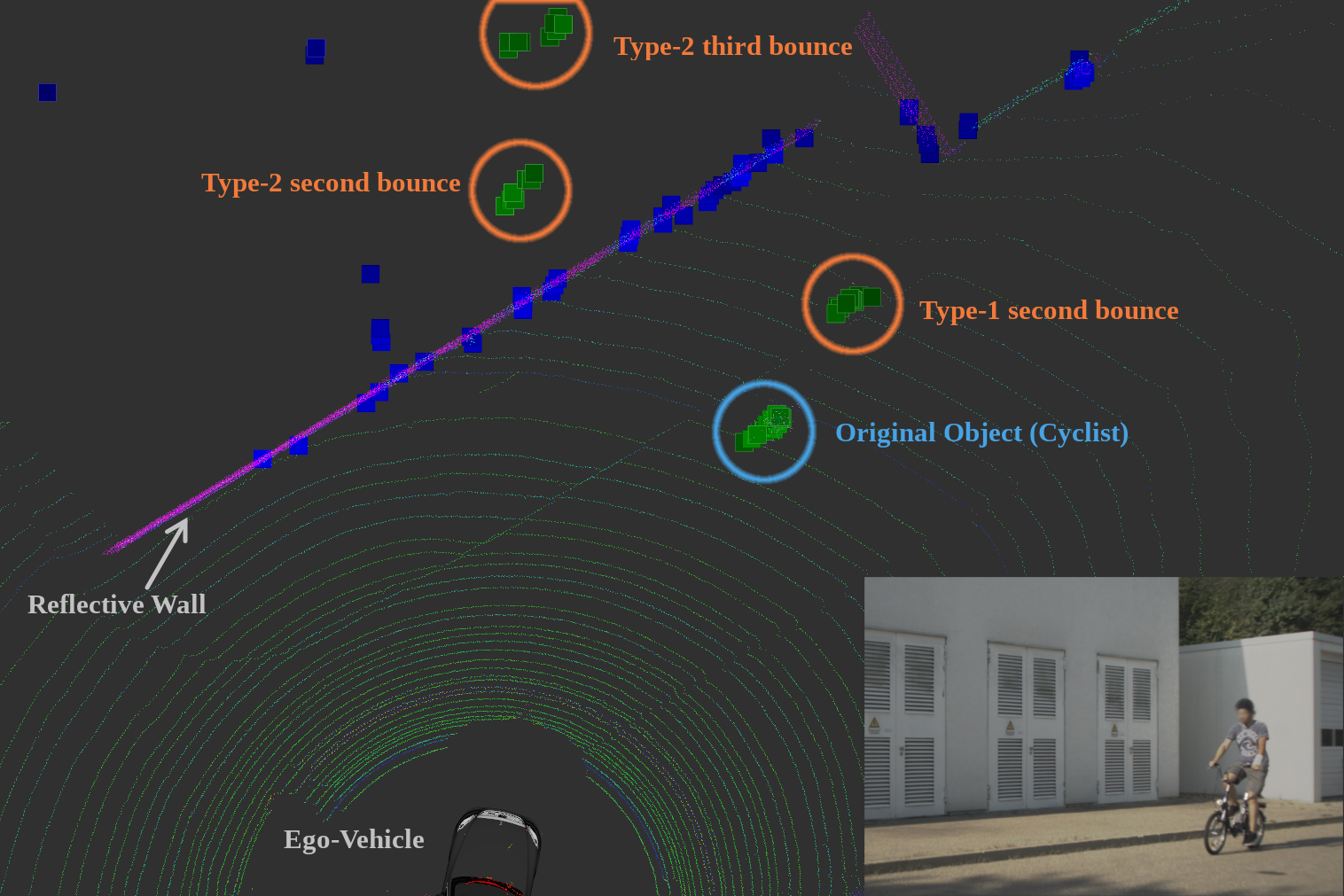}
	\caption{Real world scenario with multi-path reflections (ghost objects) of a cyclist. Different types of multi-path reflections are highlighted alongside the original ones. The small dots correspond to a lidar reference system, the bigger ones are the radar reflections. Both, radar and lidar, indicate the presence of a reflective wall which is responsible for the multi-path wave propagation.}
	\label{fig:bounce}
\end{figure}

\section{Related Work}
Multi-path radar detections are a well known phenomenon in radar processing.
They can be utilized, e.g., to detect objects via reflections beneath a car that occludes the direct path, or for height estimation with conventional 2D radar sensors \cite{laribi2018}.
Different types of multi-path occurrences are analyzed, e.g. in \cite{Kamann2018}.
Many of the possible multi-path types can be removed by using an active beam steering method for the transmission signal as well as direction-of-arrival estimation in the receive array \cite{Vermesan2013}.
However, this method requires several repeated measurements which decreases the sensor update rate.
Moreover, it cannot remove all types of multi-path effects which motivates the search for a detection system distinguishing multi-path from direct-line-of-sight objects.
To this end, other authors detect and even reconstruct ghost objects, but require knowledge about the reflector geometry \cite{sume2011radar,Qiu2014,zetik2015looking,rabaste2017around} or assume rigid objects with known motion model \cite{Roos2017} or object orientation \cite{Ryu2018}.
These assumptions can be eliminated by using a lidar system for detecting reflective surfaces as shown in \cite{Scheiner2020CVPR}.
In the latter, the ghost images are even used for a collision prevention system, highlighting the severity of mirror objects if they are undetected.
A first attempt to remove multi-path using machine learning was done by \cite{prophet2019}.
They used handcrafted features with random forests and SVM variants on a small automotive data set.

\section{Multi-Path Detections}

\begin{figure}
	\centering
	\includegraphics[width=0.99\linewidth]{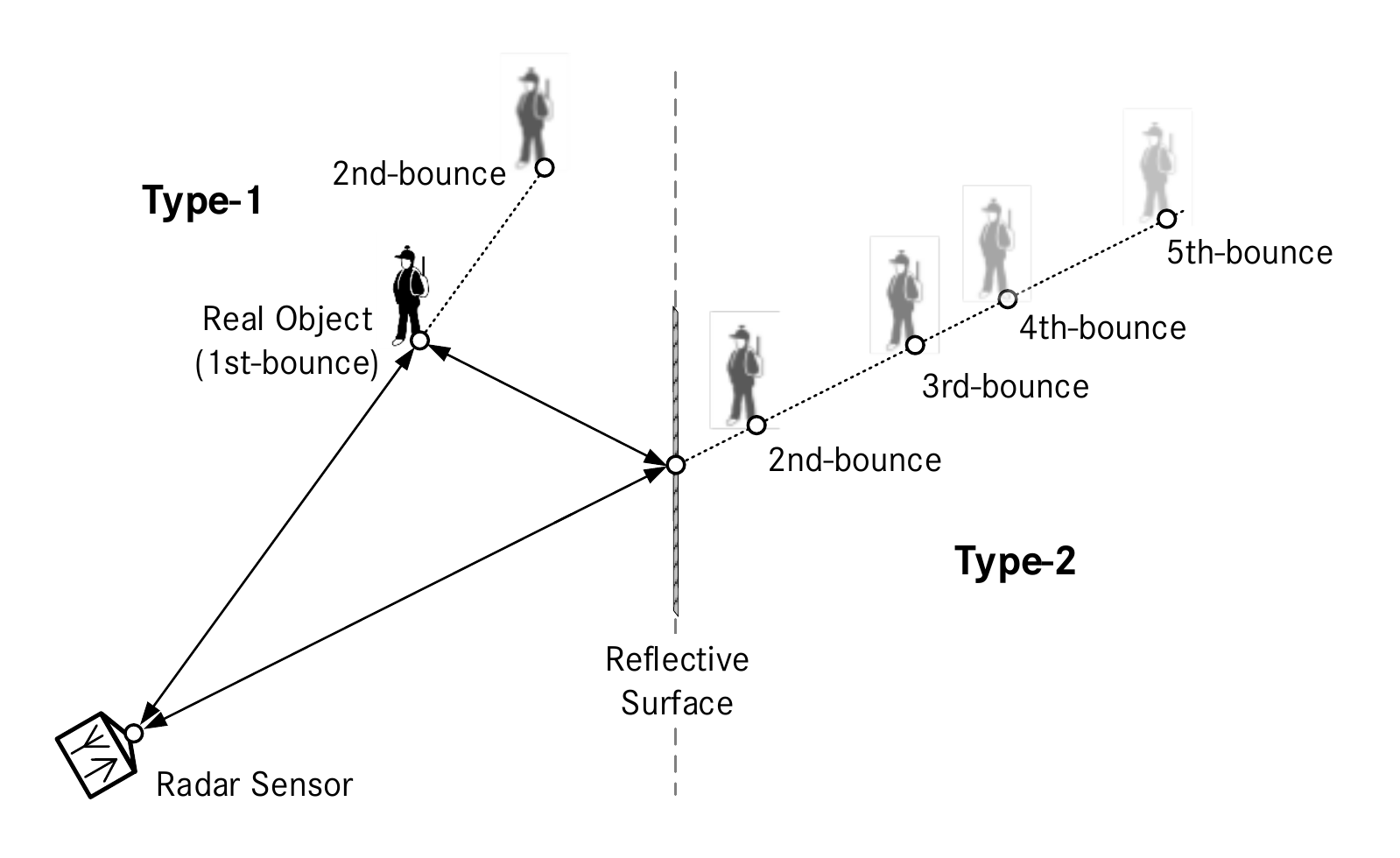}
	\caption{Illustration of different multi-path occurrences.}
	\label{fig:bounce}
	\vspace{-10pt}
\end{figure}

Ghost detections occur whenever a multi-path reflection is received by the sensor and not filtered out during early preprocessing.
For example, the signal could first bounce off a reflective surface, then to the object of interest, and back to the receiver. This example would lead to a detection at the same angle as the original object but would be perceived but in greater distance.

There are two fundamental types of multi-path reflections. Those where the last bounce happens on the object and those where it bounces off the reflective surface. To distinguish between both, we use the convention from \cite{Liu2016b} and refer to them as type-1 and type-2 reflections. Type-1 detections bounce back from the object and type-2 from the reflective surface, cf. \cref{fig:bounce}.
Furthermore, multi-path reflections are categorized by the number of objects they bounce off. A signal which is directly reflected off an object is called a first-bounce detection. Each additional bounce increases the order leading to second-, third-, and higher-order bounce detections.
With this nomenclature, the above example would be classified as a type-1 second-bounce multi-path reflection, or type-1 second-bounce detection.

Each bounce consists of a diffuse and a specular reflection part. Hence, not all energy is preserved in the signal, i.e., higher-order bounces correspond to a lower received signal energy. This leads to a rapid decrease in the amount of detection points at higher-order bounces for most reflection surfaces.
In this article, we focus on third-bounce reflections of type-2.
This focus has several reasons:
First, type-2 odd-bounce reflections can occur irrespective of the presence of a direct-line-of-sight path to the real object.
While this gives the opportunity to predict objects before they would be visible otherwise (cf. \cite{Scheiner2020CVPR}), there are no direct reflections for additional reasoning about the likelihood of the estimated detections to be ghost reflections in this case.
Second, the independence from direct-line-of-sight objects increases the amount of data that can be utilized in scenarios where the object of interest is occluded over longer periods of time, as it is case in the utilized data set.
In addition, higher-order bounces, e.g. fifth or seventh, usually result in substantially less detection points and lower signal amplitudes, i.e., they are less likely to be mistaken for a real object.

\section{Data Set}
\label{sec:dataset}
\begin{figure*}[ht]
\centering

\begin{subfigure}{.3\textwidth}
\frame{\includegraphics[width=\textwidth]{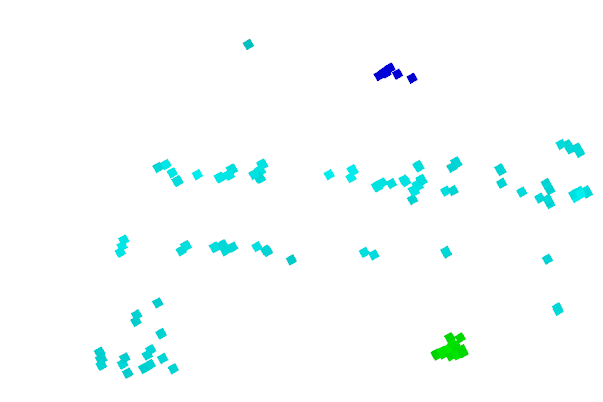}}
\caption{Model trained on: background and object. Object is correctly detected (green) and ghost is correctly classified as background (blue).~\\}
\label{fig:bg-obj-all-good}
\vspace{9px}
\end{subfigure}
\begin{subfigure}{.3\textwidth}
\frame{\includegraphics[width=\textwidth]{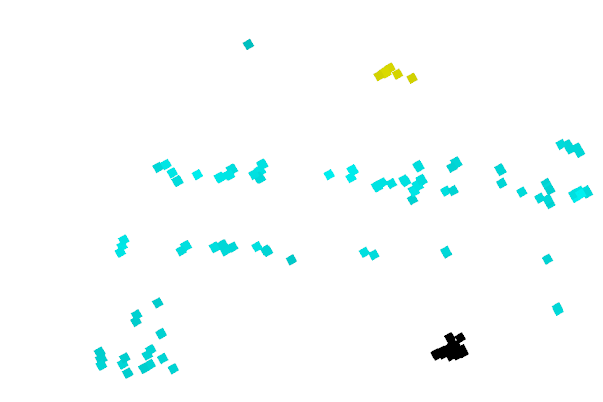}}
\caption{Model trained on: background and ghost object. Object is correctly detected as background (black) and ghost is correctly classified (yellow).}
\label{fig:bg-mirobj-all-good}
\vspace{9px}
\end{subfigure}
\begin{subfigure}{.3\textwidth}
\frame{\includegraphics[width=\textwidth]{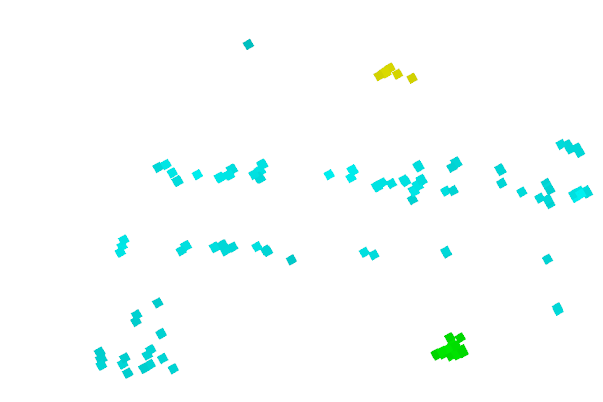}}
\caption{Model trained on: background, object, and ghost object. Object is correctly detected (green) and ghost is correctly classified (yellow).}
\label{fig:bg-obj-mirobj-all-good}
\vspace{9px}
\end{subfigure}
\begin{subfigure}{.3\textwidth}
	\frame{\includegraphics[width=\textwidth]{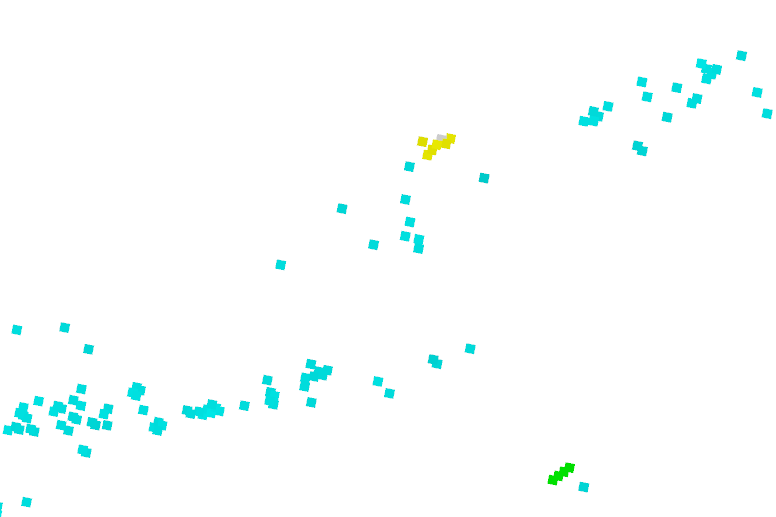}}
	\caption{Model trained on: background, object, and ghost object. All the images in this row are all one frame apart. Here everything is correctly classified.}
	\label{fig:bg-obj-mirobj-1}
	\vspace{9px}
\end{subfigure}
\begin{subfigure}{.3\textwidth}
	\frame{\includegraphics[width=\textwidth]{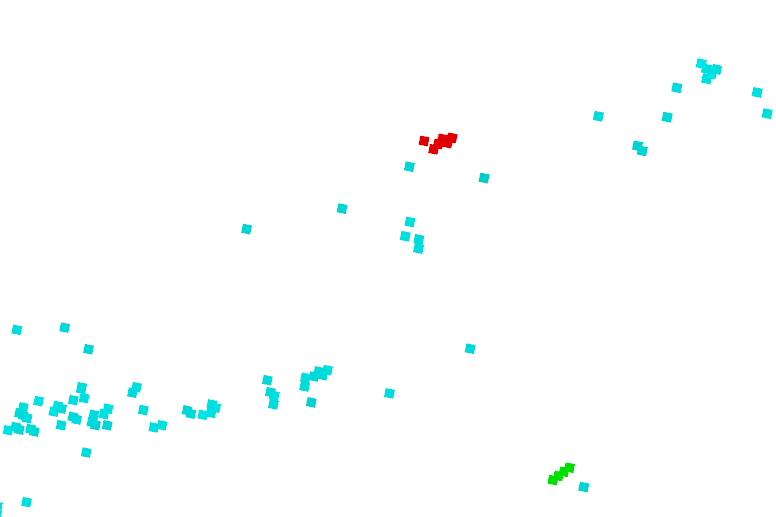}}
	\caption{In the next frame the ghost object is confused with a real object.~\\~\\}
	\label{fig:bg-obj-mirobj-2}
	\vspace{9px}
\end{subfigure}
\begin{subfigure}{.3\textwidth}
	\frame{\includegraphics[width=\textwidth]{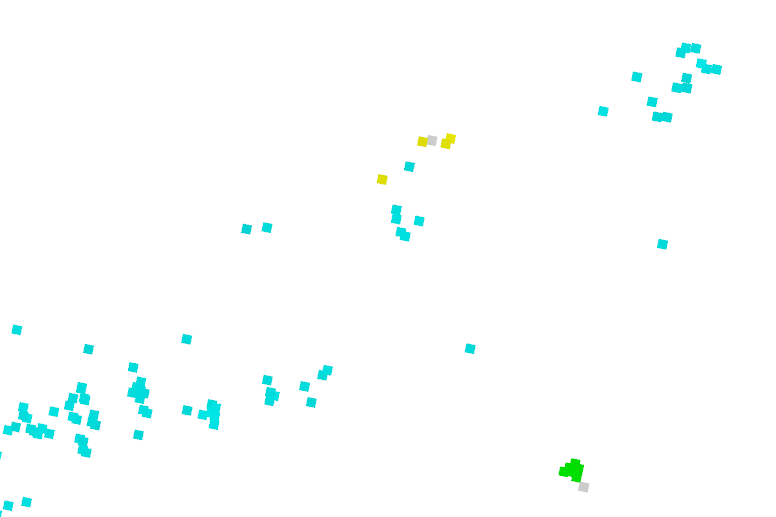}}
	\caption{Back to correctly detecting real and ghost object.~\\~\\}
	\label{fig:bg-obj-mirobj-3}
	\vspace{9px}
\end{subfigure}
\vspace{10px}
\begin{subfigure}{.3\textwidth}
	\frame{\includegraphics[width=\textwidth]{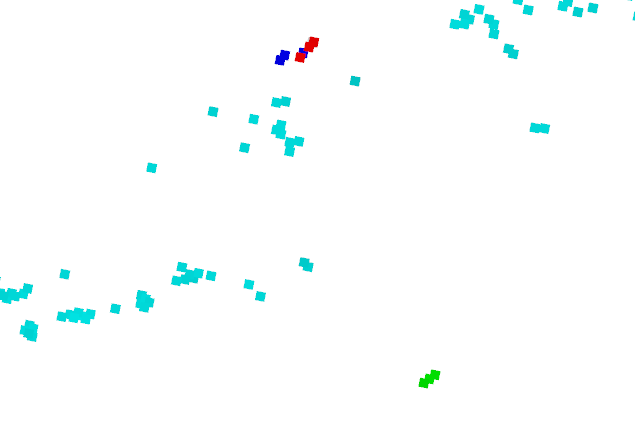}}
	\caption{Model trained on: background and object. Object is correctly detected (green) but some ghost detections are confused with a real object (red) whereas others are correctly classified as background (blue).}
	\label{fig:bg-obj-one-bad}
\end{subfigure}
\begin{subfigure}{.3\textwidth}
	\frame{\includegraphics[width=\textwidth]{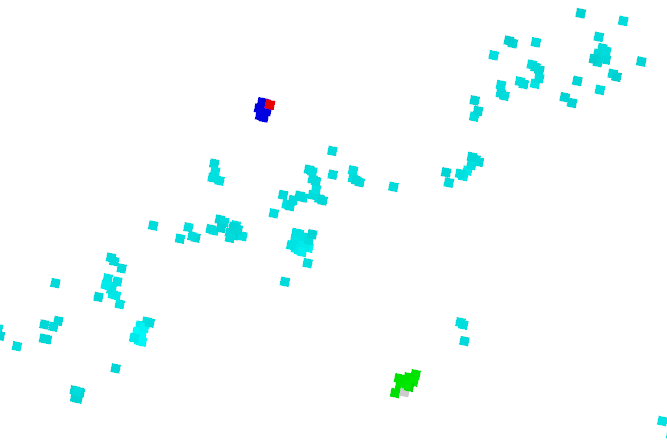}}
	\caption{Model trained on: background and object. Object is correctly detected (green) but one ghost detections is confused with a real object (red).~\\}
	\label{fig:bg-obj-multi-bad}
\end{subfigure}
\begin{subfigure}{.3\textwidth}
	\frame{\includegraphics[width=\textwidth]{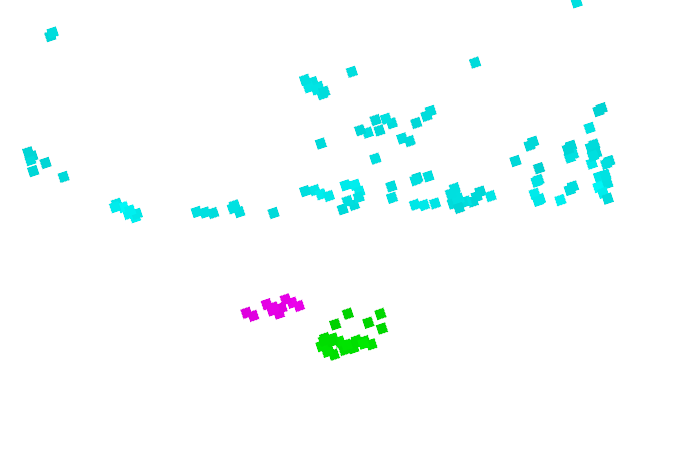}}
	\caption{Model trained on: background and object. Object is correctly detected (green) but the type-1 second-order multi-path reflections are wrongly classified as a real object (pink).~\\}
	\label{fig:bg-obj-second-bounce}
\end{subfigure}

\caption{Qualitative evaluation: The first row shows the same frame for different models all classified correctly.
	In the second row, a failure example is shown where a ghost object is wrongly classified as a real object during a single time frame.
	In the last row, different failure modes are highlighted.
	In all figures the cyan colored points represent correctly classified background.
	For better visualization we only show a small region of interest, as a result the ego-vehicle is not visible.}
\label{fig:qualitative}

\end{figure*}

For all our experiments, we use the same data set as recorded in \cite{Scheiner2020CVPR}.
The data set consists of 25 different scenarios which are repeated four times on average, resulting in a total of 100 data recordings.
Each recording contains a single vulnerable road user (VRU) which is either a pedestrian or a cyclist.
Each scenario starts with the VRU moving away from the ego-vehicle on a fixed path next to a reflector.
Reflectors comprise, e.g., parked cars, building structures, curbstones, or guardrails.
The recordings continue until the object is out of direct sight, turns around, and approaches the vehicle again.
Once the VRU is back at the ego-vehicle, the measurement stops.

For data recording, we use two experimental radar sensors mounted in the front bumper of a test vehicle.
The sensor specifications can be found in \cref{tab:sensor_specs}.
\begin{table}[tb]
	\renewcommand{\arraystretch}{1.3}
	\caption{Radar sensor specification.}
	\label{tab:sensor_specs}
	\centering
	\begin{tabular}{cccc}
		$f / \SI{}{\giga\hertz}$ & $r / \SI{}{\meter}$ & $\yaw / \deg$ & $\radvel /\SI{}{\meter\per\second}$\\
		\midrule
		$76-77$ & $0.15-153$ & $\pm70$ & $-44.3 - +44.3$ \\
		\bottomrule\\
		$\Delta_t / \SI{}{\milli\second}$ & $\Delta_r / \SI{}{\meter}$ & $\Delta_\yaw /\deg$ & $\Delta_{\radvel} /\SI{}{\meter\per\second}$\\
		\midrule
		$100$ & $0.15$ & $1.8$ & $0.087$\\
		\bottomrule
	\end{tabular}
\end{table}
The upper half of the table represents the frequency range $f$ of the emitted signal and the operational bands for range (distance) $r$, azimuth angle $\yaw$, and radial (Doppler) velocity $\radvel$ respectively.
In the second part the resolutions  $\Delta_r, \Delta_\yaw, \Delta_{\radvel}$ and $\Delta_t$ for $r$, $\yaw$, $\radvel$, and time $t$ are noted.
The radar data is labeled using a global navigation satellite system (GNSS) reference which is mounted in a wearable backpack following \cite{Scheiner2019IRS}.
All automatically labeled data were manually checked and corrected if necessary.
This step is important because high buildings, used as reflective surfaces, sometimes lead to severe multi-path errors in the GNSS signal.

Our data set originally consists of five different classes, pedestrians and cyclists, their corresponding type-2 third-bounce ghost objects, and other background detection points.
The background class consists of static points, measurement artifacts, and other clutter.

The total recording time adds to about \SI{63}{\minute}, the class distribution among detection points is given in \cref{tab:data}.
\begin{table}[tb]
	\renewcommand{\arraystretch}{1.3}
	\caption{Data set detection distribution over all classes. Ghost detections represent type-2 third-bounce reflections in this case.}
	\label{tab:data}
	\centering
	\begin{tabular}{ccccc}
		\toprule
		Pedestrian & Ghost Ped. & Bike & Ghost Cycl. & Garbage \\
		\midrule
		$2.6 \cdot 10^5$ & $3.0 \cdot 10^4$ & $2.25 \cdot 10^5$ & $2.1 \cdot 10^4$ & $2.7 \cdot 10^7$ \\
		\bottomrule
	\end{tabular}
\end{table}

\begin{figure*}[ht]
\centering

\begin{subfigure}{.3\textwidth}
\includegraphics[width=\textwidth]{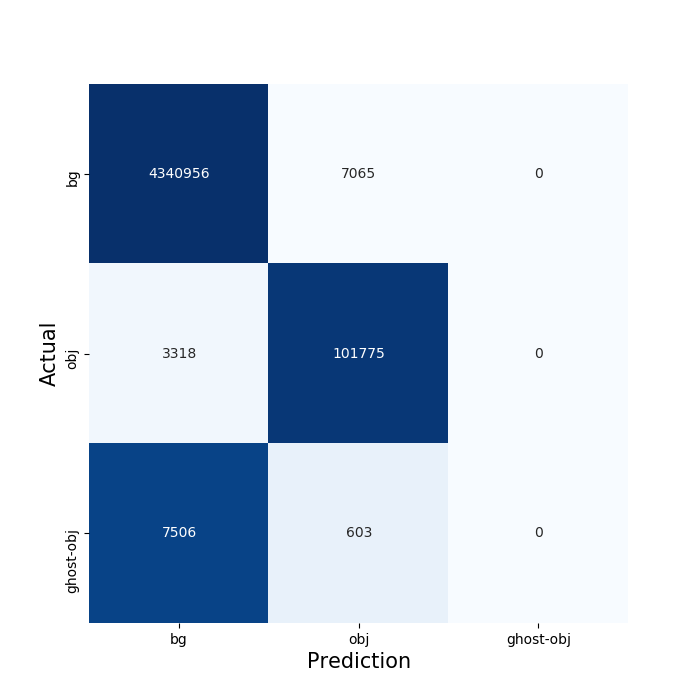}
\caption{Model trained on: background and object.~\\}
\label{fig:bg-obj-confusion}
\end{subfigure}
\begin{subfigure}{.3\textwidth}
\includegraphics[width=\textwidth]{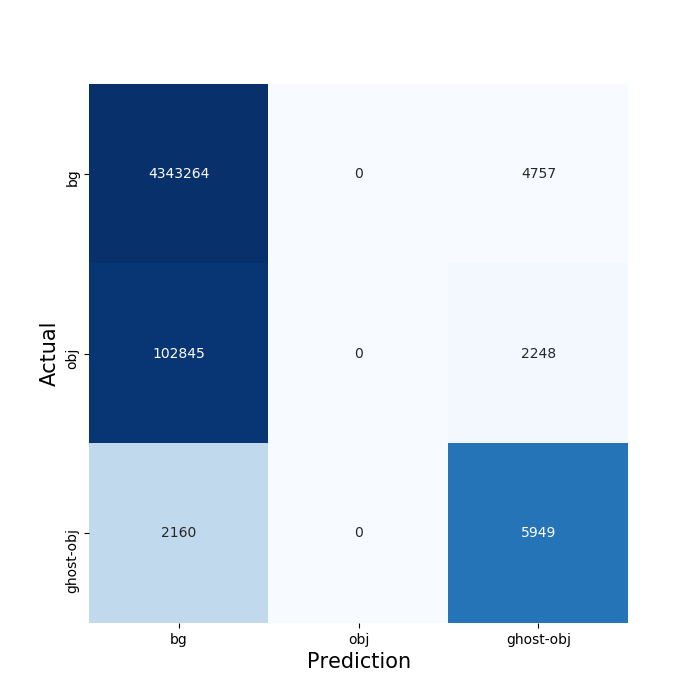}
\caption{Model trained on: background and ghost object.}
\label{fig:bg-mirobj-confusion}
\end{subfigure}
\begin{subfigure}{.3\textwidth}
\includegraphics[width=\textwidth]{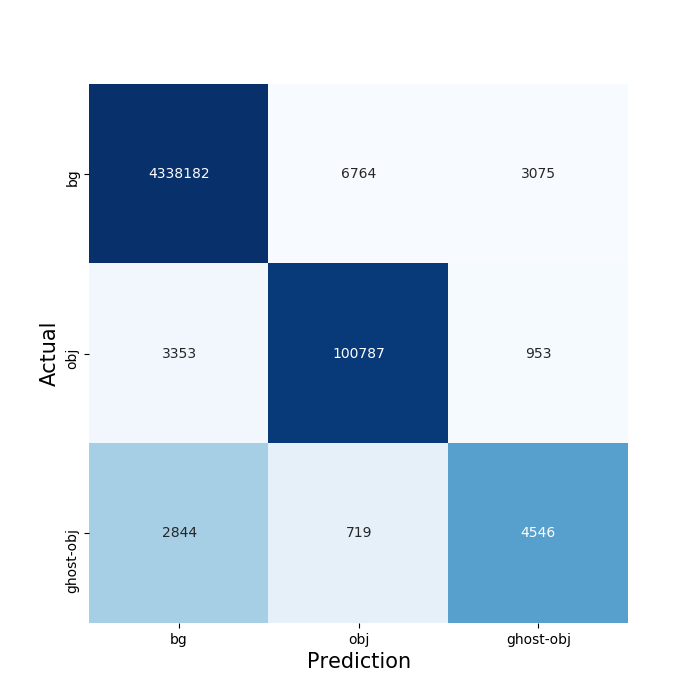}
\caption{Model trained on: background, object, and ghost object.}
\label{fig:bg-obj-mirobj-confusion}
\end{subfigure}

\begin{subfigure}{.3\textwidth}
	\includegraphics[width=\textwidth]{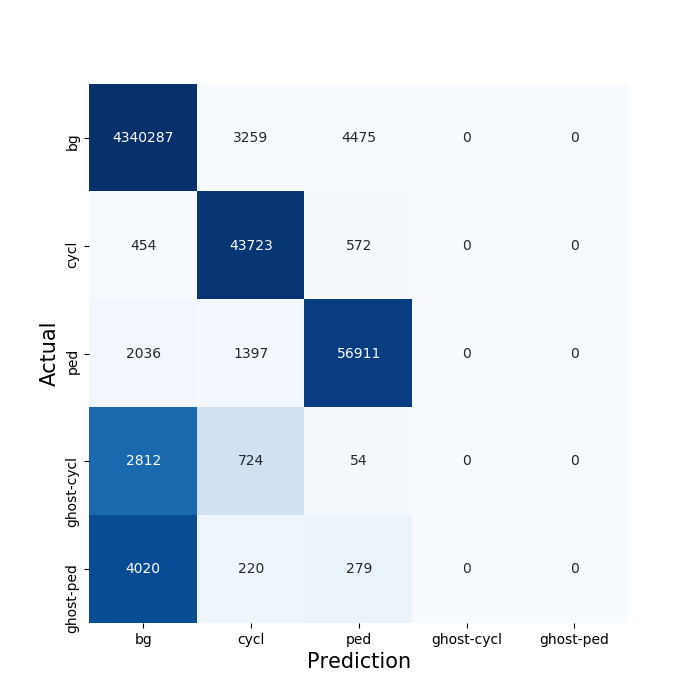}
	\caption{Model trained on: background, cyclist, and pedestrian.}
	\label{fig:bg-ped-cycl-confusion}
\end{subfigure}
\begin{subfigure}{.3\textwidth}
	\includegraphics[width=\textwidth]{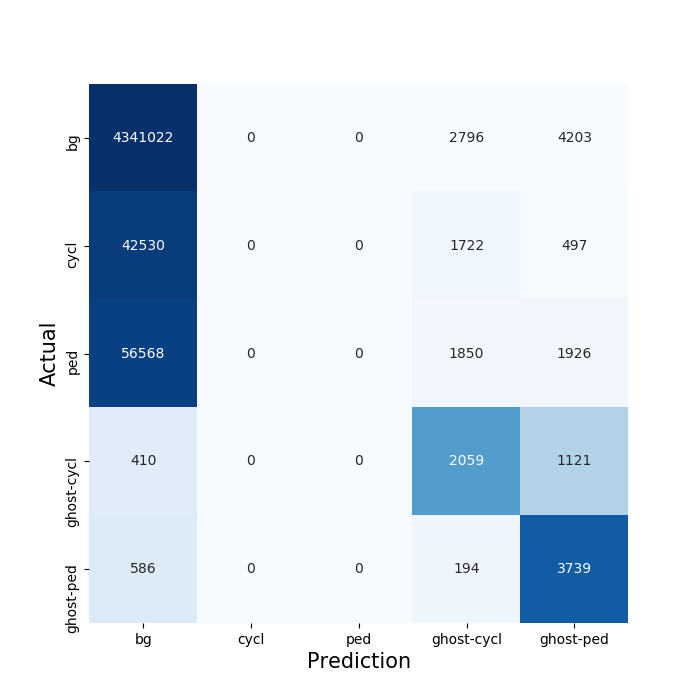}
	\caption{Model trained on: background, ghost cyclist, and ghost pedestrian.}
	\label{fig:bg-mirped-mircycl-confusion}
\end{subfigure}
\begin{subfigure}{.3\textwidth}
	\includegraphics[width=\textwidth]{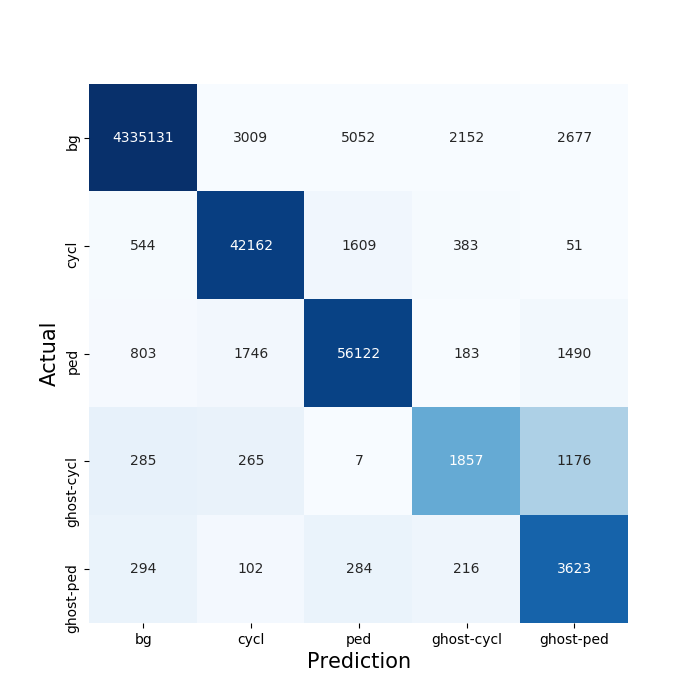}
	\caption{Model trained on: background, cycl., ped., ghost cycl., and ghost ped.}
	\label{fig:bg-ped-cycl-mirped-mircycl-confusion}
\end{subfigure}

\caption{Confusion matrices for all trained models.}
\label{fig:confusion}

\end{figure*}

\section{Methodology and Experimental Results}
Our method expects as input a radar point cloud which is resolved in range $r$, azimuth angle $\phi$, Doppler velocity $v_r$, and time $t$.
Furthermore, each point is described by an amplitude which is an estimation of the radar cross section of the reflecting object part.

\subsection{PointNet++}
To segment the radar point cloud, PointNet++ is utilized \cite{Qi2017}.
PointNet++ is an neural network architecture for processing point cloud data (PCD).
It is based on the earlier PointNet architecture which was developed to consume point clouds without preprocessing steps such as rendering the point cloud into an image grid.
PointNet++ can be used for classification and semantic segmentation of point clouds.

When used for segmentation tasks the architecture of PointNet++ resembles those of hour-glass networks used for image based semantic segmentation.

The point cloud is downsampled multiple times into a coarser representation by using "multi scale grouping" (MSG) modules.
Each MSG module consists of three stages. First, iterative farthest point sampling is applied to collect center points.
Second, clusters are build around those center points by grouping all points within a epsilon region around the center point.
This second grouping stage is applied for different radii, thus the name "multi scale grouping".
Then, in a third and final step, each cluster or local region is processed by a mini PointNet to encode local features.
The output of the final stage equals the new input to the next MSG module.

In a second upsampling stage, the downsampled data is upsampled to reconstruct the original point cloud and provide a label for each original data point.
The upsampling is handled by "Feature Propagation" (FP) modules. These modules propagate the low-level features to the original points.

After the upsampling, the features are processed by fully connected (FC) layers with dropout for a final classification of each point.

Since PointNet++ was designed to process 3D data, the original code is changed to handle the 2D data provided by radar sensors, as demonstrated in \cite{Schumann2018}.

For more information on PointNet++ see the original paper \cite{Qi2017} and \cite{Schumann2018} where it was first used to segment radar data.

The architecture proposed by \cite{Schumann2018} is used.
It consists of three MSG modules with three corresponding FP modules and three subsequent FC layers.
The input point cloud is fixed to 2048 points.
We accumulate the radar data over a time window of \SI{200}{\milli\second}, this is a common approach to counteract the sparsity of radar data.
Furthermore this is an easy and effective way to combine the data of our dual-sensor setup.
In \cite{Schumann2018}, they accumulated the data over \SI{500}{\milli\second} due to a sparser radar system than the one used by us.
If the accumulated point cloud does not consist of at least 2048 points, the ones with the highest amplitude are duplicated.
If there are too many points, then the points with the lowest amplitude are discarded.

To account for class imbalances present in the data set, the loss for each point is scaled relative to the inverse proportion each class takes up in the data set.
This results in the following loss $L$:

\begin{equation}
L(p, l) = \frac{1}{c\cdot s_l} \cdot \cross-entropy(p, l)
\end{equation}
with the ground truth $l$, the prediction $p$, the total number of classes $c$ and the proportion $s_l$ of the occurrence of $l$ in the data set.

For each experiment, the network is trained for 100 epochs and the best checkpoint is selected based on the performance on the validation split.
A random subset of $10\%$ from the training set is used as a validation set.

\subsection{Experimental Setup}
\label{subsec:experiments}
\begin{table}[tb]
	\caption{Training Setups}
	\label{tab:experiments}
	\centering
	\begin{tabular}{rl}
		\toprule
		Index & Classes used during training         \\ \midrule
		1 & bg, obj                              \\
		2 & bg, ghost-obj                        \\
		3 & bg, obj, ghost-obj                   \\ \midrule
		4 & bg, ped, cycl                        \\
		5 & bg, ghost-ped, ghost-cycl            \\
		6 & bg, ped, cycl, ghost-ped, ghost-cycl \\ \bottomrule
	\end{tabular}
\end{table}
\begin{table*}[tb]
    \caption{Scores for all experiments.}
    \label{tab:data}
    \centering
    \begin{tabular}{@{}rccccccccc@{}}
    	        \toprule
    	    \textbf{Score} & \textbf{Obj} & \textbf{Ghost-Obj} & \textbf{Ped} & \textbf{Ghost-Ped} & \textbf{Cycl} & \textbf{Ghost-Cycl} & \textbf{Average}\setfootnotemark & \textbf{Bg} & \textbf{Experiment} \\ \midrule
    	       \textbf{F1} &   $94.88$    &                    &              &                    &               &                     &     $94.88$      &   $99.87$   &          1          \\
    	                   &              &      $56.49$       &              &                    &               &                     &     $56.49$      &   $99.90$   &          2          \\
    	                   &   $94.47$    &      $54.50$       &              &                    &               &                     &     $74.49$      &   $99.82$   &          3          \\
    	\cmidrule(r){2-10} &              &                    &   $92.81$    &                    &    $92.96$    &                     &     $92.89$      &   $99.87$   &          4          \\
    	                   &              &                    &              &      $46.72$       &               &       $33.72$       &     $40.22$      &   $99.84$   &          5          \\
    	                   &              &                    &   $90.95$    &      $53.53$       &    $91.62$    &       $44.31$       &     $70.10$      &   $99.83$   &          6          \\ \midrule
    	   \textbf{Recall} &   $96.84$    &                    &              &                    &               &                     &     $96.84$      &   $99.82$   &          1          \\
    	                   &              &      $73.36$       &              &                    &               &                     &     $73.36$      &   $99.84$   &          2          \\
    	                   &   $95.90$    &      $56.06$       &              &                    &               &                     &     $75.98$      &   $99.77$   &          3          \\
    	\cmidrule(r){2-10} &              &                    &   $94.31$    &                    &    $97.71$    &                     &     $96.01$      &   $99.79$   &          4          \\
    	                   &              &                    &              &      $82.74$       &               &       $57.35$       &     $70.05$      &   $99.71$   &          5          \\
    	                   &              &                    &   $93.00$    &      $80.17$       &    $94.22$    &       $51.73$       &     $79.78$      &   $99.70$   &          6          \\ \midrule
    	\textbf{Precision} &   $92.99$    &                    &              &                    &               &                     &     $92.99$      &   $99.92$   &          1          \\
    	                   &              &      $45.92$       &              &                    &               &                     &     $45.92$      &   $99.95$   &          2          \\
    	                   &   $93.09$    &      $53.02$       &              &                    &               &                     &     $73.05$      &   $99.86$   &          3          \\
    	\cmidrule(r){2-10} &              &                    &   $91.36$    &                    &    $88.65$    &                     &     $90.00$      &   $99.94$   &          4          \\
    	                   &              &                    &              &      $32.55$       &               &       $23.88$       &     $28.22$      &   $99.98$   &          5          \\
    	                   &              &                    &   $88.98$    &      $40.18$       &    $89.17$    &       $38.76$       &     $64.27$      &   $99.96$   &          6          \\ \bottomrule
    \end{tabular}
\end{table*}

For training, the labels are split into two experiments.
One experiment with the original five labels described in \cref{sec:dataset}: \emph{Pedestrians} (ped), \emph{cyclists} (cycl), \emph{ghost pedestrians} (ghost-ped), \emph{ghost cyclists} (ghost-cycl), and \emph{background} (bg).
The second experiment groups (ghost) pedestrians and (ghost) cyclists together in the (ghost) \emph{object} class (obj, ghost-obj).
The first experiment aims not only to classify vulnerable road users against the background but also to discriminate between different kinds of road users.
In the second experiment, the aim is to separate relevant objects from the background without further differentiation.

For each experiment, three models are trained. One standard model which only classifies objects (ped, cycl), another which classifies only ghost objects, and a third one classifying ghost and real objects. An overview is given in \cref{tab:experiments}.

When evaluating each model, all classes from the associated experiments are used to create a full confusion matrix.

During evaluation, all respective classes are used, e.g., the model trained on background vs. object was evaluated against background, object, and ghost object.
In this manner, it is possible, to evaluate the confusion of each model between ghost and real objects.

\section{Results}
In this section, the quantitative and qualitative results of the experiments described in \cref{subsec:experiments} are presented.

\subsection{Quantitative Evaluation}
\subsubsection{Performance Metrics}
For the evaluation, we employ three different scores: precision, recall, and F1. Where the latter is the harmonic mean between recall and precision. We employ those scores for each class separately. The scores are defined as follows. Precision is the percentage of correct predictions per class:
\begin{equation}
\text{precision} = \frac{\text{true positives}}{\text{true positives} + \text{false positives}}.
\end{equation}
Recall is the percentage of correctly identified points per class:
\begin{equation}
\text{recall} = \frac{\text{true positives}}{\text{true positives} + \text{false negatives}}.
\end{equation}
The F1 score combines the above into a single score as the harmonic mean of precision and recall:
\begin{align}
\text{F1} = 2\cdot \frac{\text{precision} \cdot \text{recall}}{\text{precision} + \text{recall}}.
\end{align}

\subsubsection{Evaluation}
The results for this evaluation are listed in \cref{tab:data}. Furthermore, confusion matrices are presented in \cref{fig:confusion}.

First, we compare the difference in performance when training only on objects, ghosts, or both at once.
We note a slightly higher F1 score for the object class when only training on real objects compared to training on real objects and ghosts.
However, the difference of $94.88\%$ to $94.47\%$ is minor, highlighting that it is possible to detect ghost objects without sacrificing performance on real objects.
When only training on ghost objects vs. background, we detect an increase of $2\%$ from $54.50\%$ to $56.49\%$ in the F1 score on ghost objects compared to the model jointly trained on ghosts and objects.
However, this increase in the F1 score is due to an large increase in recall ($56.06\%$ to $73.36\%$).
At the same time, a decrease in precision is present ($53.02\%$ to $45.92\%$).
The only slightly increased F1 score reflects this loss in precision.
\footnotetext[1]{Average in table does not include background (bg) score, since only the scores on the foreground classes are relevant for the overall (average) model performance.}  

This is further demonstrated by the two confusion matrices in \cref{fig:bg-mirobj-confusion} and \cref{fig:bg-obj-mirobj-confusion}.
The model trained on ghosts and background correctly classifies $5949$ ghost detections, but also misclassifies $2248$ real objects ($4757$ background points) as ghosts. The model trained on objects, ghosts, and background, however, only misclassifies $953$ ($3075$) points while only correctly classifying $4546$ ghosts.

Based on those results, it seems beneficial to train combined on objects and ghost objects at the same time to reduce confusion between the two.

When evaluating the model only trained on object vs. background, another interesting fact is gained from those experiments.
Only $603$ out of $7668$ false positives are due to misclassification of a real object as a ghost, accounting for only $7.86\%$ of all false positives, cf. \cref{fig:bg-obj-confusion}.
This is certainly less than expected, showcasing that modern deep-learning-based approaches already can distinguish between real and ghost objects.
This percentage is slightly increased to $9.61\%$ when jointly training on objects and ghost objects.
However, in a qualitative analysis we find that misclassified ghosts are often correctly classified as ghosts a few timesteps earlier.
This is further discussed in \cref{subsec:qualitative}.

A similar picture is painted when looking at the models trained on pedestrians and cyclists separately. However, the performance difference when jointly training on ghosts and real objects versus only training on real objects is slightly larger.

The most interesting fact occurs when comparing the model trained on background, pedestrians, and cyclists to one trained on background and object.
A much larger proportion of false positives are due to confusion with ghost objects, cf. \cref{fig:bg-ped-cycl-confusion}. For the cyclist class $944$ out of $5600$ false positives are due to misclassified ghosts or $16.86\%$.
If we ignore the false positives on the pedestrian class, this increases to $22.46\%$.
On the other hand, for the pedestrian class we only have $6.19\%$ or $6.93\%$.
This suggests that when trying to discriminate between pedestrians and cyclists, the model has a higher chance to get confused by ghost objects.
This is also highlighted by the fact that the model trained on pedestrians and cyclists has a total of $1268$ false positives on ghosts compared to the $603$ of the model trained on background vs. object.
This is an increase of roughly $100\%$. Whereas, the false positives on the background only increases by roughly $10\%$: $7734$ compared to $7064$.

Nevertheless, this effect vanishes completely when jointly training on background, pedestrians, cyclists, ghost pedestrians, and ghost cyclists, cf. \cref{fig:bg-ped-cycl-mirped-mircycl-confusion}.
In this scenario only $658$ false positives are due to ghost objects, an improvement of almost $100\%$.

But on the other hand, this yields a higher confusion between real pedestrians and cyclists.

From this we conclude, it is beneficial to add ghost labels when training on multiple classes.
Even if this increases intra-class confusion it greatly reduces the amount of false positives due to ghost objects.
In either case adding ghost objects to the training data yields a respectable F1 score of around $55\%$ for ghost objects.
Although not on par with the detection rate of real objects this can be beneficial in scenarios where ghost objects are desired as, e.g., in \cite{Scheiner2020CVPR} or when trying to estimate the existance and orientation of a mirror-like surface.

\subsection{Qualitative Examples}
\label{subsec:qualitative}
In \cref{fig:qualitative}, qualitative results are shown.
The first row consists of images where all models correctly segment the scene according to their different training objectives.

In the second row, an interesting case is highlighted.
Each image in this row is one frame apart from the other.
In the first and the last image the ghost object is correctly classified as such (\cref{fig:bg-obj-mirobj-1}, \cref{fig:bg-obj-mirobj-3}).
However, in the middle frame the ghost object is suddenly misclassified as a real object (\cref{fig:bg-obj-mirobj-2}).
This shows that using temporal information could be of great value when combined with our work.
For example, applying a tracker could help to suppress those types of failure modes.

In the last row, two confusions of ghost objects with mirrors are showcased, cf. \cref{fig:bg-obj-one-bad} and \cref{fig:bg-obj-multi-bad}.
In \cref{fig:bg-obj-second-bounce}, we showcase a failure mode of our training data.
The model wrongly classifies a type-1 second-bounce object as a real object.
These second-bounce objects are not annotated in our data set and, thus, are easily misclassified as real objects.

\section{Discussion}
In this article, we evaluate the effects of ghost objects on modern deep-learning-based approaches on a large-scale automotive data set.
Furthermore, we show that by using labeled ghost objects during training, we can detect those challenging objects with a precision and recall of around $53\%$ and $56\%$, respectively.
We found that adding ghost labels to the training scenarios with multiple positive labels, e.g., pedestrians and cyclists, dramatically reduces false positives.
There are also certain shortcomings to our approach, all due to missing data.
In a qualitative analysis we found false positives due to unlabeled second-bounce multi-path reflections.
Those kinds of reflections are currently not annotated in our data set.
Labeling those other multi-path reflections is challenging task and requires a lot of resources.
Annotating even the real first-bounce reflection is hard, because of sparse and noisy radar data, therefore, we used an auto-labeling system for our annotations.
Due to the largely unknown reflection characteristics of the VRUs in our experiments, this approach is not feasible for second-order multi-path reflections.
The insights of this article indicate the potential of our approach and database.
It shows that modern machine learning algorithms can indeed distinguish real objects from ghost detections.
For future work, we plan to annotate all type-1 and type-2 reflections manually which will allow for a better analysis of ghost objects in radar data and to evaluate more sophisticated temporal approaches such as tracking.
Furthermore, it would be interesting to investigate ghost objects caused by other road users such as cars or motorcycles.

\section*{Acknowledgment}
The research leading to these results has received funding from the European Union under the H2020 ECSEL Programme as part of the DENSE project, contract number 692449.

\bibliographystyle{IEEEtran}
\bibliography{newbib}

\end{document}